\PassOptionsToPackage{table, x11names}{xcolor}
\documentclass[10pt,twocolumn,letterpaper]{article}

\usepackage{iccv}              % To produce the CAMERA-READY version
\usepackage{multirow}
\definecolor{iccvblue}{rgb}{0.21,0.49,0.74}
\usepackage[pagebackref,breaklinks,colorlinks,allcolors=iccvblue]{hyperref}
\usepackage[accsupp]{axessibility}

\def\paperID{4507} % *** Enter the Paper ID here
\def\confName{ICCV}
\def\confYear{2025}

\title{Exploiting Domain Properties in Language-Driven Domain Generalization for Semantic Segmentation}

\author{Seogkyu Jeon$^{1}$ \quad Kibeom Hong$^{2*}$ \quad Hyeran Byun$^{1}\thanks{Corresponding author}$\vspace{0.2cm}\\
$^{1}$Yonsei University\quad\quad $^{2}$Sookmyung Women’s University\\
{\tt\small jone9312@yonsei.ac.kr}
}

\begin{document}
\maketitle
\begin{abstract}

% The recent advance of the large scale vision-language models (VLMs) have greatly enhanced the power of the downstream vision tasks, especially semantic segmentation. However, they still struggle from unexpected domain shift between the source and the target domain which leads to performance degradation in the real-world scenario.

Recent domain generalized semantic segmentation (DGSS) studies have achieved notable improvements by distilling semantic knowledge from Vision-Language Models (VLMs). However, they overlook the semantic misalignment between visual and textual contexts, which arises due to the rigidity of a fixed context prompt learned on a single source domain. To this end, we present a novel domain generalization framework for semantic segmentation, namely Domain-aware Prompt-driven Masked Transformer (DPMFormer). Firstly, we introduce domain-aware prompt learning to facilitate semantic alignment between visual and textual cues. To capture various domain-specific properties with a single source dataset, we propose domain-aware contrastive learning along with the texture perturbation that diversifies the observable domains. Lastly, to establish a framework resilient against diverse environmental changes, we have proposed the domain-robust consistency learning which guides the model to minimize discrepancies of prediction from original and the augmented images. Through experiments and analyses, we demonstrate the superiority of the proposed framework, which establishes a new state-of-the-art on various DGSS benchmarks. The code is available at \url{https://github.com/jone1222/DPMFormer}.
\end{abstract}    
\section{Introduction}
\label{sec:intro}
Over the decades, semantic segmentation has made remarkable progress, now being able to precisely classify each pixel in an image into categories. However, one of the shadows that lies in these advancements is that models often exhibit inconsistent and degraded performance when deployed in various real-world environments.
% To overcome the discrepancy between the training and test domains, \ie, domain shift, the studies of domain generalization~\cite{gulrajani2021search, cheng2024disentangled, yu2024rethinking, addepalli2024leveraging} have drawn attention which aims to derive domain-robust knowledge from the source domain that is generalizable on various target domains.
% Domain Generalized Semantic Segmentation (DGSS) has progressed through diverse approaches, such as feature whitening~\cite{pan2018two, pan2019switchable, choi2021robustnet, peng2022semantic} and domain randomization~\cite{yue2019domain, huang2021fsdr, zhao2022style, lee2022wildnet, wu2022siamdoge, zhong2022adversarial, fan2023towards, jiang2023domain, chattopadhyay2023pasta, kim2023texture}. 
To this end, the task of Domain Generalized Semantic Segmentation (DGSS) has arisen to overcome the discrepancy between the training and test domains, \ie, domain shift. DGSS has progressed through diverse approaches, such as feature whitening~\cite{pan2018two, pan2019switchable, choi2021robustnet, peng2022semantic} and domain randomization~\cite{yue2019domain, huang2021fsdr, zhao2022style, lee2022wildnet, wu2022siamdoge, zhong2022adversarial, fan2023towards, jiang2023domain, chattopadhyay2023pasta, kim2023texture}. Furthermore, some studies~\cite{ding2023hgformer, bi2024learning} exploited multi-scale object queries with a transformer-based architecture, \ie, Mask2Former~\cite{cheng2022masked}.
% Moreover, advanced framework, \ie Mask2former

\begin{figure}[t]
  \centering
  \includegraphics[clip=true,width=0.95\linewidth]{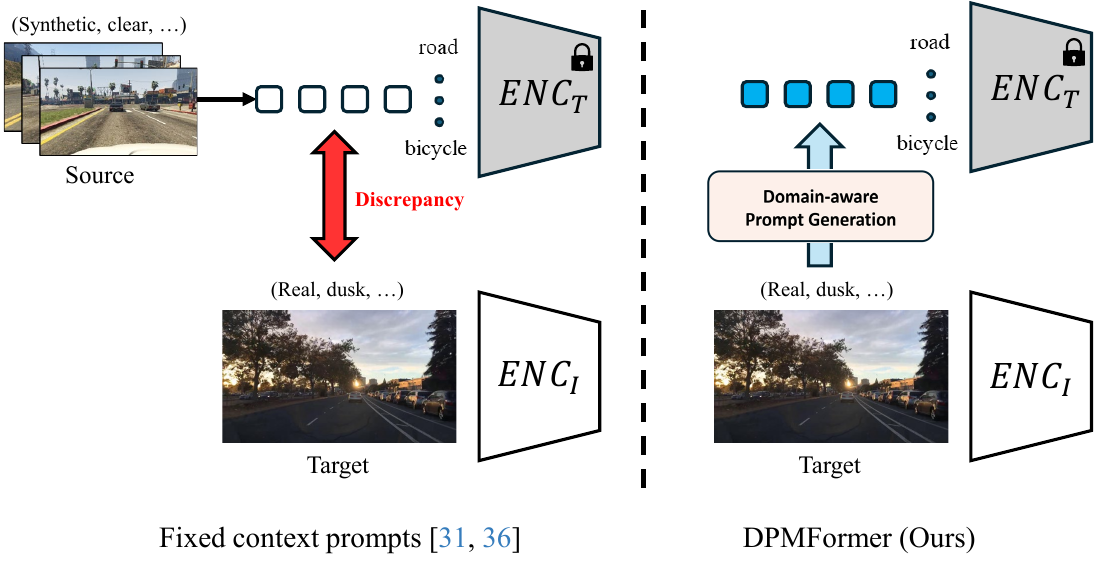}
  \caption{Motivation of DPMFormer.  Using a fixed context prompts~\cite{lin2023clip, pak2024textual} tend to retain source domain properties, causing contextual misalignment with the target domain. On the other hand, DPMFormer translates domain properties of the input image into context prompts, enhancing semantic alignments.}
  \label{fig:motivation}
\end{figure}

\begin{figure*}[t]
  \centering
  \includegraphics[clip=true,width=0.90\linewidth]{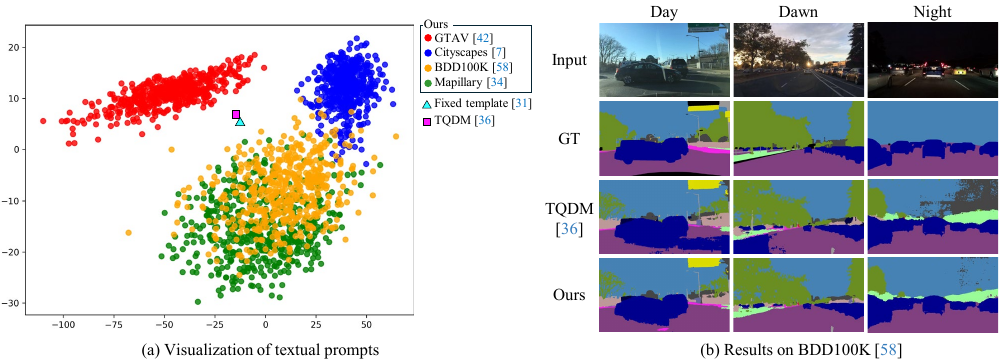}
  \caption{PCA visualization of textual prompts (left) and qualitative results on various environments (\eg, Day, Dawn, Night) in BDD100K~\cite{yu2020bdd100k} (right). The models are trained on GTAV~\cite{richter2016playing} with the CLIP-pretrained backbone (ViT-B)~\cite{radford2021learning}. A fixed single-context prompt lacks flexibility in adapting to various domain shifts due to its rigidity. In contrast, our framework utilizes domain-specific properties from input images as context prompts, enhancing semantic alignment between text and images. As a result, as shown in (b), our approach exhibits improved robustness across diverse environments.}
  \label{fig:intro}
\end{figure*}

Despite these advancements, learning domain-robust representations solely from the single source domain remains a significant hurdle to performance improvement. Recently, several studies~\cite{fahes2024simple,wei2024stronger,hummer2023vltseg,pak2024textual} employed Vision-Language Models (VLMs)~\cite{radford2021learning, sun2023eva} owing to their semantic knowledge learned from diverse large-scale text-image datasets. For this, pioneering VLM-based DGSS works~\cite{hummer2023vltseg, fahes2024simple, wei2024stronger} have adopted their pre-trained visual encoder for initialization and fine-tuning~\cite{hummer2023vltseg, wei2024stronger}. 
In addition, TQDM~\cite{pak2024textual} has introduced a framework that constructs object queries from the textual descriptions, leveraging semantic concepts from linguistic expressions.

Yet, they still have limitations in fully utilizing textual knowledge to improve domain generalizability. Concretely, the context prompts that decorates textual descriptions of each category are either a predefined template~\cite{lin2023clip} (e.g., '\textit{a photo of}') or a single learnable text embedding~\cite{pak2024textual}. However, we contend that fixed context prompts have limited capability for target domain images due to the discrepancy between visual and textual contexts as depicted in Fig.~\ref{fig:motivation}. Firstly, a handcrafted template inherently encodes the characteristics of a specific domain which restricts its generalizability to other unseen domains. Secondly, prompt optimization is prone to source domain overfitting, especially in the single-source setting. Consequently, both approaches may enlarge the gap between textual and visual contexts, leading to suboptimal performances on target domains. 

% should include details like its darkened exterior or the reflections of city lights.
In this perspective, we point out that the semantic form of the textual representation for an object category should be changed with respect to the input visual context in order to strengthen their semantic correspondence. For instance, in the real-world driving scene at night of Fig.~\ref{fig:intro} (b), target categories (\eg, \textit{car} and \textit{sky}) possess distinct textures from those of synthetic daytime images. Hence, it would be more appropriate to decorate \textit{`a car'} with the textual prompt \textit{`at night in the real-world'} which is reflecting the domain characteristics. As a result, the modified text feature will include details such as darkened car exteriors and light reflections, enhancing the semantic alignment. 
Nevertheless, it is challenging to design domain-adaptive prompts in the DGSS setting where the training dataset covers only a single domain. Moreover, this setting hinders the visual pipeline from learning robust image features against domain shifts.

To address these challenges, we introduce a novel framework, namely Domain-aware Prompt-driven Masked Transformer (DPMFormer), focusing on two key aspects: (1) Leveraging domain-specific properties of the input image (domain-awareness) and (2) Generating accurate outputs on images with dissimilar domain characteristics (domain-robustness). To cultivate the domain-awareness, we propose a novel \textbf{domain-aware prompt learning}, which translates the domain-specific properties of an input image into context prompts via an auxiliary network. In addition, to obtain diverse domain properties in the single-source setting, we apply texture perturbations to synthesize novel domain images. Furthermore, exploiting both source and novel domain images, we propose a \textbf{domain-aware contrastive loss} to ensure that the derived prompts effectively capture domain-specific properties of the input image. This loss encourages the anchor context prompts to be distinguishable from those of different domain characteristics, while being closer to those from the anchor domain. With the proposed domain-aware prompt, the model accurately identifies target classes while being aware of the input domain.
% These images exhibit different domain characteristics from the original ones, they provide useful features for deriving domain-specific properties. 

% However, since the training dataset solely comes from a source domain, the auxiliary network may overfit to the training domain, leading to poor generalization on unseen domains.
% To prevent the prompt generation network from source domain overfitting, we generate novel domain images through random photometric transformations. By using images from the source and generated domains, we encourage the output prompt to be distinguishable between images from different domains while being similar to the output of its original domain, using domain separability loss and domain grouping loss, respectively. Derived prompts are then integrated with category-specific textual prompts, allowing the trained model can adaptively identify target in aware of the domain-specific property of the input image during inference.

% Since we initialize the model with a pretrained VLM that has been trained on a large scale dataset, the model already has the ability to recognize various domains. If we can provide a hint about which domain the target image belongs to, the model should be available to leverage its learned knowledge for the task. For example, in a night driving scene, the model will detect the car more accurately when the textual query is specified as 'a photo of a car at night' rather than simply 'a photo of a car'.

Moreover, we strive to provide better domain robustness guidances to the visual encoder and decoders. We carefully organize the texture perturbations with structure-preserving image transformations, ensuring that the original visual context remains intact. Instead of simply reusing the original ground truths for novel domain images, we introduce \textbf{domain-robust consistency loss} to guarantee reliable predictions under severe domain shifts. The loss consists of class consistency and mask consistency losses, which penalize discrepancies between class and mask predictions of given image pairs, respectively. Furthermore, domain-robust consistency losses are applied at every layer of the transformer decoder, preventing discrepancies in earlier layers from propagating to later parts. 
% especially in mask predictions, which are used for attention masking.

% Rather than simply reusing the original ground-truth to learn categories from novel domains with augmented images, we guide the model to generate consistent predictions even after textual changes.

Through evaluations on various DGSS benchmarks, we demonstrate the superiority of the proposed framework. Notably, our framework achieves state-of-the-art semantic segmentation performance across domain generalization scenarios. Additionally, ablation studies and detailed analyses validate the effectiveness of each component. 

\section{Related Works}
\label{sec:related_work}

\subsection{Domain Generalized Semantic Segmentation}
Domain Generalized Semantic Segmentation (DGSS) aims to learn domain-invariant representations that generalize robustly to various unseen target domains. The task assumes a single-source setting, where only one dataset is available during training. Unlike domain adaptation~\cite{csurka2017comprehensive, zou2018unsupervised, tsai2018learning} and test-time domain adaptation~\cite{wang2020tent, yang2021generalized, wang2022continual}, access to target domains is strictly prohibited, making DGSS more challenging.
% The domains in the training dataset are denoted as "source domains" whereas the domains of the test dataset are indicated as "target domains".
Previous studies approached DGSS in two ways primarily: feature whitening and normalization~\cite{pan2018two, pan2019switchable, choi2021robustnet, peng2022semantic}, and domain randomization approaches~\cite{pan2018two, yue2019domain, pan2019switchable, huang2021fsdr, zhao2022style, lee2022wildnet, wu2022siamdoge, zhong2022adversarial, fan2023towards, jiang2023domain, chattopadhyay2023pasta, kim2023texture}. 

Feature whitening and normalization approaches~\cite{pan2018two, pan2019switchable, choi2021robustnet, peng2022semantic} mainly focus on removing features which are variant to domain shifts. Exploiting the characteristics of instance whitening~\cite{li2017universal} and instance normalization~\cite{ulyanov2017improved} which can effectively remove texture and style from the input image, these approach leverages those operations in between the backbone module to minimize the effect from domain and texture changes. Representatively, RobustNet~\cite{choi2021robustnet} introduced instance selective whitening module that finds gram matrix components sensitive to photometric changes and minimize their changes. However, due to the natural difficulty in disentangling domain-specific and domain-invariant features, these approaches show limited performance gains.

Domain randomization approaches~\cite{yue2019domain, huang2021fsdr, zhao2022style, lee2022wildnet, wu2022siamdoge, zhong2022adversarial, fan2023towards, jiang2023domain, chattopadhyay2023pasta, kim2023texture} augment novel domains from the source domain by modifying either the images or their features through various methods, \eg affine transformations~\cite{huang2017arbitrary, zhao2022style, lee2022wildnet, wu2022siamdoge, zhong2022adversarial, fan2023towards}, image translation~\cite{zhu2017unpaired, yue2019domain}, frequency decomposition~\cite{huang2021fsdr, chattopadhyay2023pasta},and photometric transformations~\cite{jiang2023domain}. Synthesized samples increase the domain diversity of the training dataset, reducing the domain gap between the learned representation and the test data. Moreover, most of the generative DGSS methods adopt context-preserving transformations to compute the output discrepancy between the original sample and the augmented one. Representatively, SHADE~\cite{zhao2022style} proposed a style consistency loss to encourage model to learn invariant pixel-level semantic information by minimizing the Jenson-Shannon Divergence (JSD) between the output predictions of original and augmented images. 

Meanwhile, several studies~\cite{ding2023hgformer, bi2024learning} have built DGSS upon Mask2Former~\cite{cheng2022masked}, that leverages attention mechanism~\cite{vaswani2017attention} renowned for robustness against domain shifts~\cite{hendrycks2019benchmarking, szegedy2013intriguing, hendrycks2021many}. Mask2former~\cite{cheng2022masked} involves a transformer decoder that exploits object query features to group pixels of same objects or categories. Based on this, HGFormer~\cite{ding2023hgformer} first proposed a hierarchical framework that groups pixels to form part-level masks for complementing whole-level pixel grouping procedure. Similarly, CMFormer~\cite{bi2024learning} utilizes down-sampled features additionally via feature fusion which are more domain-invariant than the original features.

\subsection{Language-driven Domain Generalized Semantic Segmentation}

Recent studies~\cite{fahes2024simple,wei2024stronger,hummer2023vltseg,pak2024textual, huo2024domain} has investigated to exploit Vision Language Models (VLMs)~\cite{desai2021virtex, jia2021scaling, lu2019vilbert, pham2023combined, radford2021learning, tan2019lxmert, sun2023eva} for DGSS, owing to its powerful generalizability learned from large-scale datasets of image-text pairs. VLTseg~\cite{hummer2023vltseg} employs the image encoder of VLMs for initializing backbone parameters, and fine-tune the network with the segmentation objective function. FAMix~\cite{fahes2024simple} proposes to yield class-specific novel styles by concatenating random style description and class names, then performs style randomization by locally mixing the source and the novel styles. DAP~\cite{huo2024domain} exploits text encoder to distill the semantic textual knowledge of target categories to the visual backbone. TQDM~\cite{pak2024textual} points out deficient use of language information in aforementioned works, and introduces a textual-query driven framework for DGSS. They utilize the textual description of each categories for initializing input query features for transformer decoder, and also for computing the text-to-pixel attention to enhance semantic clarity of pixel features. Despite their advancements, we argue that two points are overlooked : (1) the domain discrepancy between the learned textual prompt and the domain-specific property of the input image (2) lack of domain robustness guidance. To mitigate these limitations, we carefully refine the textual queries with the captured domain-specific property of the input image, and augment novel domain images to improve domain-awareness of the model as well as prediction consistency learning.
% Despite their advancements, we argue that two points are overlooked : (1) the importance of textual prompt design in aware of the domain-specific property of the input image (2) lack of domain robustness guidance. To mitigate these limitations, we carefully refine the textual queries with the captured domain-specific property of the input image, and augment novel domain images to improve domain-awareness of the model as well as prediction consistency learning.
% Textual prompts of each categories should include not only class-specific information but also domain-specific information of the input image to achieve better alignment between the text and image, fully leveraging the generalizability of VLMs. Also, training the whole framework only on a single training domain have deficiency in improving domain robustness while being prone to source domain overfitting. 
\begin{figure*}[t]
  \centering
  \includegraphics[clip=true,width=\linewidth]{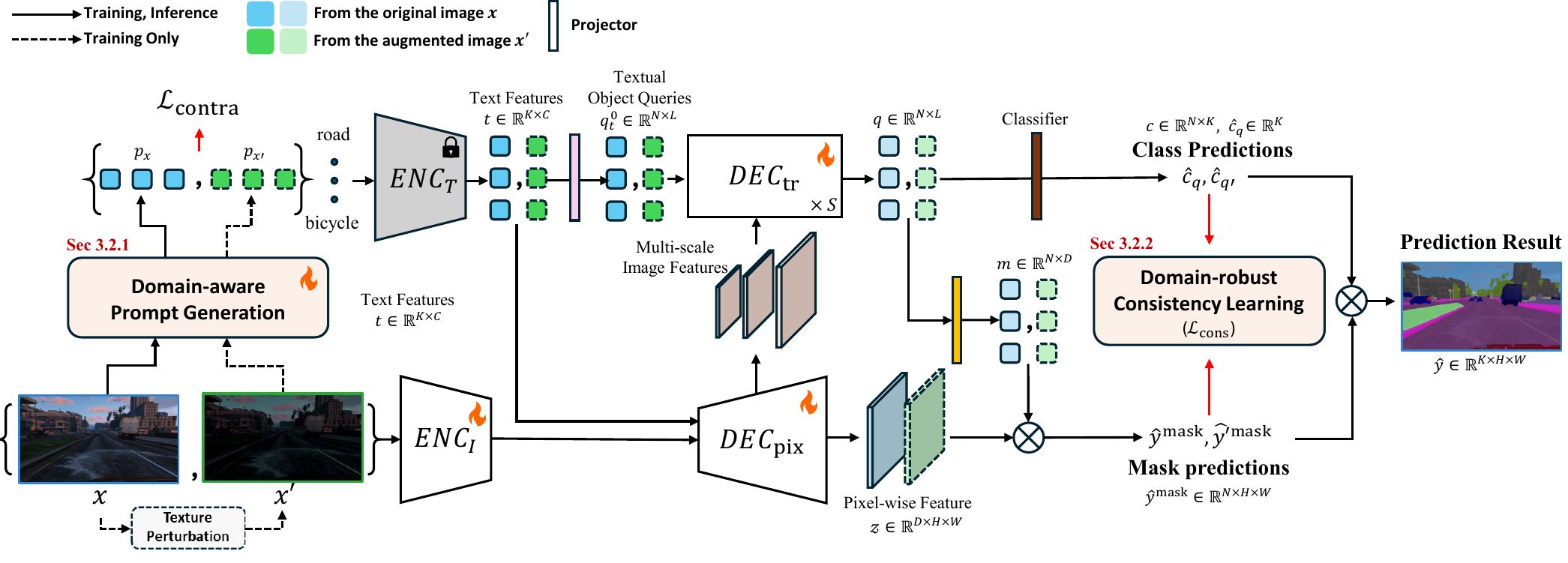}
  \caption{Illustration of DPMFormer. We use Mask2Former~\cite{cheng2022masked} based architecture which consists of a backbone image encoder ($ENC_{I}$), a pixel decoder ($DEC_{\text{pix}}$), a transformer decoder ($DEC_{\text{tr}}$), and a text encoder ($ENC_{T})$. During training, we synthesize images with a novel domain style via texture perturbation. Both images are incorporated to compose a batch and exploited for learning domain-awareness (Sec.~\ref{sec:domain_awareness}) and domain-robustness (Sec.~\ref{sec:domain_robustness}). }
  \label{fig:architecture}
\end{figure*}

\section{Methods}
%이제 기초는 다 썼고, 우리 설계가 어떤 novelty가 있는지 강조하는 거 주말동안 추가해야함

\subsection{Preliminaries}

We design our DPMFormer based on mask classification architecture, \ie, Mask2Former~\cite{cheng2022masked}. Mask2Former consists of an image encoder $\textit{ENC}_{I}$, pixel decoder $\textit{DEC}_{\text{pix}}$, and transformer decoder $\textit{DEC}_{\text{tr}}$. The input RGB image $x \in \mathbb{R}^{3 \times H \times W}$ is first fed to the image encoder for feature extraction, and then converted to pixel-wise features $z \in \mathbb{R}^{D \times H \times W}$ through the pixel decoder. The transformer decoder iteratively refines $N$ object queries $q\in\mathbb{R}^{N \times L}$ with multi-scale image features from the pixel decoder. Thereafter, refined object queries are projected into mask embeddings $m \in \mathbb{R}^{N \times D}$ to generate pixel-level mask prediction via dot product as $\hat{y}^{\text{mask}} = m \cdot z, \hat{y}^{\text{mask}} \in \mathbb{R}^{N \times H \times W}$. The category label of each object query $q$ is predicted with a linear classifier, \ie, $c_{q} \in \mathbb{R}^{K}, c \in \mathbb{R}^{N \times K}$. The final prediction result is derived by the matrix multiplication between the pixel-level mask prediction and class prediction of queries as $\hat{y} = c^  {\top} \cdot \hat{y}^{\text{mask}}, \hat{y} \in \mathbb{R}^{K \times H \times W}$.

Furthermore, inspired by \cite{pak2024textual}, we employ a text encoder $\textit{ENC}_{T}$ for initializing object queries from textual descriptions. Pre-trained vision language mdoels (VLMs), \eg, CLIP~\cite{radford2021learning}, is exploited to initialize both $\textit{ENC}_{I}$ and $\textit{ENC}_{T}$ parameters, and the parameters of $\textit{ENC}_{T}$ remain frozen during training to preserve learned linguistic knowledge. The text encoder $\textit{ENC}_{T}$ produces a text feature $t_{k} \in \mathbb{R}^{C}$ corresponding to each class label $\text{class}_{k}$ with a learnable context prompt $p$, \ie, $t_{k} = {ENC}_{T}([p, \{\text{class}_{k}\}])$. Then, obtained text embeddings $t=\{t_{k}\}^{K}_{k=1}\in\mathbb{R}^{K\times{C}}$ are passed to a multi-layer perceptron (MLP) to derive initial textual object queries $q^{0}_{t}$.  Also, the text-to-pixel attention mechanism is established in a pixel decoder layer to enhance pixel semantic clarity enhancement. 
% Therefore, the number of queries $N$ equals to the number of classes $K$ in this setting.

% In addition, three regularization losses are introduced to maintain alignment between visual and textual features, which is formulated as $\mathcal{L}_{reg} = \mathcal{L}_{reg}^{L} + \mathcal{L}_{reg}^{V} + \mathcal{L}_{reg}^{VL}$. The overall training loss for TQDM is set as $\mathcal{L}_{base} = \mathcal{L}_{seg} + \mathcal{L}_{reg}$.

\subsection{DPMFormer}
As illustrated in Fig.~\ref{fig:architecture}, Domain-aware Prompt-driven Masked Transformer (DPMFormer), aims to cultivate two core aspects of domain generalization for the model: domain-awareness and domain-robustness. For domain-awareness, we propose the domain-aware context prompt learning which translates domain-specific property of the input image into the text embedding for semantic alignment between textual and visual features. And for domain-robustness, we encourage the model to generate accurate outputs against textural changes via consistency learning. %of the image via consistency learning.

% \noindent\textbf{Texture Augmentation} Since DGSS assumes the training dataset consists of a single source domain, it is challenging to learn diverse domain properties without any additional data. 
% Based on the intuition that domain generalizability is related to the number of observable domain in the training dataset, we stylize the training dataset to generate an auxiliary domain dataset. 

\noindent\textbf{Texture perturbation.} In order to effectively guide the model with diverse domain characteristics, we stylize the training dataset to generate an auxiliary domain dataset. Following RobustNet~\cite{choi2021robustnet}, we adopt photometric transformations—comprising strong color jittering, gaussian blur and noise injection—for their simplicity as well as content-preserving property. The generated image $x'$ is combined with its original image $x$ to form a batch for training.
% offering two benefits (1) As the transformed image has different textural features from the original source domain, the model can be exposed to diverse domain characteristics during training, enabling it to effectively handle novel domains. (2) By using content-preserving transformations, the original semantics and learned knowledge can be further leveraged to enhance the prediction consistency of the model. 

%\subsubsection{Learning domain-awareness via domain-specific context prompt}
\subsubsection{Domain-aware context prompt learning}
\label{sec:domain_awareness}
% \subsection{Improving domain-awareness }
% Textual query generation plays a crucial role in enabling the model to leverage the linguistic semantics of each class previously learned by the VLM.
% We employ prompt learning~\cite{zhou2022learning} to adapt text embeddings for textual query generation and semantic segmentation task. 
Textual query generation and prompt learning~\cite{zhou2022learning} plays a crucial role in enabling the model to leverage the linguistic semantics of each class previously learned by the VLM. Although the learned context prompt is beneficial for performance improvement, it is optimized solely within the source domain without direct consideration of domain shifts. This single context prompt may yield strong results on several domain shift scenarios where target domain has similar domain characteristics with the source. However, in cases of severe domain shift, the performance gain may be limited due to the contextual mismatch between the target domain image and the learned prompt. For example, if the prompt learning is conducted on a dataset containing only sunny day images, the learned prompt is could be misaligned when encountering rainy night scenes. As such, the semantic misalignment should be addressed in order to fully utilize rich semantic knowledge of pre-trained VLM. To this end, we propose to generate domain-aware context prompt that extracts domain-specific properties from the input image as a context prompt.

% , inspired by CoCoOp~\cite{zhou2022conditional},
To obtain domain-aware context prompt from the input image $x$, we design an auxiliary network $h_\theta(\cdot)$ that takes visual feature as an input and generates a domain-specific prompt embedding $\pi_x = h_\theta(\hat{F}(x))$, where $\hat{F}(x)$ denotes a visual feature extracted from a frozen visual backbone of CLIP~\cite{radford2021learning}. We use the class token as a visual feature $\hat{F}(x)$ for its global representation~\cite{dosovitskiy2020vit}. Nextly, the obtained domain-specific prompt embedding $\pi_x$ is integrated with the context prompt embedding $p$ through addition, \ie, $p_x = p + \pi_x$, then concatenated with text embeddings of classes to generate domain-aware textual features as $t_{x, k} = {ENC}_{T}([p_x, \{\text{class}_{k}\}])$. To encourage derived textual features to include domain-specific information of the input image, we introduce a novel domain-aware contrastive learning framework with the original and augmented images. The loss function is depicted as follows:
\begin{equation}
\label{equ:contra}
    \mathcal{L}_{contra} = - \frac{1}{2B} \sum_{i=1}^{2B} \log \frac{ \sum_{j \in \mathcal{P}_i} \exp \text{sim}(\pi_{i}, \pi_{j}) / \tau} { \sum_{j \in \mathcal{P}_i \cup \mathcal{N}_i} \exp \text{sim}(\pi_{i}, \pi_{j}) / \tau},
\end{equation}
where $\text{sim}(\cdot)$ means similarity metric and $\tau$ is a temperature parameter. $B$ denotes the batch size of original images, $\mathcal{P}_i$ and $\mathcal{N}_i$ are positive sets and negative sets of the $i$-th image, respectively. The positive set $\mathcal{P}_i$ is composed of the indices of samples having same domain characteristics with the anchor $i$ whereas the others belong to negative set $\mathcal{N}_i$. For example, when an anchor $i$ is of original source domain images, other original source domain images belongs to $\mathcal{P}_i$ while all augmented images are included in $\mathcal{N}_i$. In case of an augmented image as an anchor, $\mathcal{P}_i=\{i\}$ and $\mathcal{N}_i = \{1, \ldots, 2B\} \setminus i$. With the proposed loss, we encourage $h_{\theta}$ to capture domain-specific property of the image and reflect it to the output text feature $t_{k}$. Also, the final domain-aware text feature is guided by the task loss $\mathcal{L}_{seg}$ to be aligned with the original image and to minimize segmentation errors. We note that the proposed loss considers the domain information unlike CoCoOp~\cite{zhou2022conditional} which treats $\mathcal{P}_i$ as the text feature corresponding to a specific image feature while the remaining images forming the negative set. We provide comparative analysis with CoCoOp in Sec.\ref{ablation_cocop}.
\subsubsection{Domain-robust consistency learning}
\label{sec:domain_robustness}

% Thanks to the semantic preserving texture augmentation, we can effectively reuse the ground truths of the original image to train with the augmented images. Therefore, the task loss can be formulated as follows:

% \begin{equation}
% \begin{split}
% \mathcal{L}_{\text{task}} &= \lambda_{\text{mask}} \cdot (\mathcal{L}_{\text{mask}}(\hat{m}_{Q}, \mathcal{M}) + \mathcal{L}_{\text{mask}}(\hat{m}_{Q'}, \mathcal{M})) \\
% &\quad + \lambda_{\text{cls}} \cdot (\mathcal{L}_{\text{cls}}(\hat{c}_{Q}, \mathcal{C}) + \mathcal{L}_{\text{cls}}(\hat{c}_{Q'}, \mathcal{C}))
% \end{split}
% \end{equation}
% where Q and Q' denotes a set of queries of the original and augmented image, respectively. $\mathcal{M}$ and $\mathcal{C}$ are a set of ground truth masks and classes respectively, which are adequately assigned with DAHM. Following Mask2Former, $\mathcal{L}_{\text{mask}}$ is computed as a weighted sum of binary cross entropy loss and dice loss. Since the model is forced to produce predictions close to the ground truth for augmented images, it learns to make accurate predictions across various domain changes during training.

To further enhance the domain robustness, we encourage model to generate persistent predictions in the domain shift scenario. To enhance prediction consistency, we induce the model to minimize the prediction discrepancy in terms of the mask and class label as follows:
\begin{equation}
\mathcal{L}_{\text{cons}} = \sum_{s=1}^{S} \lambda_{\text{mc}} \cdot \mathcal{L}_{\text{mc}}(\hat{y}^{\text{mask}}_{\text{s}}, \hat{y'}^{\text{mask}}_{\text{s}}) + \lambda_{\text{cc}} \cdot \mathcal{L}_{\text{cc}}(\hat{c}_{q_{i}, \text{s}}, \hat{c}_{q'_{i}, \text{s}}).
\end{equation}
$S$ denotes the number of transformer blocks in the transformer decoder, and $\mathcal{L}_{\text{mc}}$ and $\mathcal{L}_{\text{cc}}$ represent the mask and class consistency losses, respectively. To compute the loss at the $s$-th transformer decoder block, we obtain $\{\hat{y}^{\text{mask}}_{\text{s}}, \hat{y'}^{\text{mask}}_{\text{s}}\}$ and $\{\hat{c}_{q_{i}, \text{s}}, \hat{c}_{q'_{i}, \text{s}}\}$ which are pairs of mask predictions and class predictions of $i$-th query $q_{i}$ from the original and augmented image pair $\{x, x'\}$, respectively. We employ binary cross entropy and Jensen–Shannon divergence as a discrepancy measure of the mask consistency loss ($\mathcal{L}_{mc}$) and class consistency loss ($\mathcal{L}_{cc}$), respectively. With the help of domain-aware context prompts and above losses, our model learns to predict not only accurately but also consistently in various domain shift scenarios during training.

\subsection{Overall Loss Functions}

The overall loss of our framework is a weighted sum of the task loss $\mathcal{L}_{\text{seg}}$, VLM regularization loss $\mathcal{L}_{\text{reg}}$, domain-aware contrastive loss $\mathcal{L}_{\text{contra}}$, and the consistency loss $\mathcal{L}_{\text{cons}}$.
\begin{equation}
\mathcal{L}_{\text{total}} = \mathcal{L}_{\text{seg}} + \lambda_{\text{reg}}\mathcal{L}_{\text{reg}} + \lambda_{\text{contra}}\mathcal{L}_{\text{contra}} + \lambda_{\text{cons}}\mathcal{L}_{\text{cons}},
\end{equation}
where $\{\lambda_{\text{reg}}, \lambda_{\text{contra}}, \lambda_{\text{cons}}\}$ are constant weighting factors. We note that $\mathcal{L}_{\text{seg}}$ and $\mathcal{L}_{\text{cons}}$ are calculated with all queries and its predictions from every block of the transformer decoder. We provide details of the baseline losses ($\mathcal{L}_{\text{seg}}$, $\mathcal{L}_{\text{reg}}$) in the supplementary.
\section{Experiments}

\subsection{Implementation Details}

\noindent \textbf{Datasets.} 
We validate DPMFormer on \textit{synthetic-to-real} and \textit{real-to-real} scenarios in the single-source setting.

\noindent \textbf{Synthetic datasets.} GTAV~\cite{richter2016playing} is a representative synthetic dataset which consists of 24,966 images with a resolution of 1914$\times$1052. The training split contains 12,403 images, while the validation and test set are of 6,382 and 6,181 images, respectively. SYNTHIA~\cite{ros2016synthia} dataset provides 6,580 images for training and 2,820 images for validation respectively, with the image resolution at 1280$\times$760. 

\noindent \textbf{Real-world datasets.} Cityscapes~\cite{cordts2016cityscapes} is a dataset collected from the real environment. The resolution of each image is 2048$\times$1024, and the population of training and validation split is 2,975 and 500, respectively. BDD100K~\cite{yu2020bdd100k} includes 7,000 training images and 1,000 validation images with the resolution of 1280$\times$720. Mapillary~\cite{neuhold2017mapillary} is composed of images with diverse resolution, where the training and validation size is 18,000 and 2,000, respectively.

\noindent \textbf{Network architecture.} %The network is 
Our approach leverages the vision transformer-based models as backbones, initialized with either the CLIP \cite{radford2021learning} (ViT-B) or EVA02-CLIP \cite{sun2023eva} (EVA02-L) model. The CLIP backbone is configured with a patch size of 16, while the EVA02-CLIP backbone uses a patch size of 14. For the pixel and transformer decoder, aforementioned, we employ a mask classification architecture~\cite{cheng2022masked}, consisting of $N=9$ layers with masked attention mechanisms.
Additionally, we design the auxiliary network $h_\theta$ for domain-aware context prompt generation as a shallow multi-layer structure (BatchNorm-Linear-ReLU-Linear).

\noindent \textbf{Training and evaluation.} To optimize DPMFormer, we employ an AdamW~\cite{loshchilov2017decoupled} where the learning rate is set as $1\times10^{-5}$ and $1\times10^{-4}$ for synthetic and real training datasets, respectively. Following previous transformer-based studies~\cite{pak2024textual, hoyer2022daformer}, we apply linear warm-up~\cite{goyal2017accurate} for initial 1,500 iterations and rare class sampling~\cite{hoyer2022daformer}. The optimizer settings are identical for both CLIP and EVA02-CLIP backbone models. We set the total training iterations and the batch size as 20,000 and 8, respectively. Weighting factors $\{\lambda_{\text{reg}}, \lambda_{\text{contra}}, \lambda_{\text{cons}}\}$ are set as {1, 1, 10}. We crop the input image to have a resolution of 512$\times$512. The texture perturbation is only applied during training. We use mean Intersection over Union (mIoU)~\cite{everingham2015pascal} to quantitatively evaluate the results following the convention.

\begin{table}[t]
    \centering
    \resizebox{1.0\linewidth}{!}{
    \begin{tabular}{l l c c c c}
        \toprule
        \textbf{Models (GTAV)} & \textbf{Backbone} & \textbf{Cityscapes} & \textbf{BDD} & \textbf{Mapillary} & \textbf{Avg.} \\
        \midrule
        SAN-SAW~\cite{peng2022semantic} & ResNet-101 & 45.33 & 41.18 & 40.77 & 42.43 \\
        WildNet~\cite{lee2022wildnet} & ResNet-101 & 45.79 & 41.73 & 47.08 & 44.87 \\
        SHADE~\cite{zhao2022style} & ResNet-101 & 46.66 & 43.66 & 45.50 & 45.27 \\
        TLDR~\cite{kim2023texture} & ResNet-101 & 47.58 & 44.88 & 48.80 & 47.09 \\
        FAMix*~\cite{fahes2024simple} & ResNet-101 & 49.47 & 46.40 & 51.97 & 49.28 \\
        \midrule
        SHADE~\cite{zhao2022style} & MiT-B5 & 53.27 & 48.19 & 54.99 & 52.15 \\
        IBAFormer~\cite{sun2023ibaformer} & MiT-B5 & 56.34 & 49.76 & 58.26 & 54.79 \\
        VLTSeg*~\cite{hummer2023vltseg} & ViT-B & 47.50 & 45.70 & 54.30 & 49.17 \\
        TQDM*~\cite{pak2024textual} & ViT-B & \underline{57.50} & \underline{47.66} & \underline{59.76} & \underline{54.97} \\
        \rowcolor{gray!40} DPMFormer* (ours) & ViT-B & \textbf{59.00} & \textbf{51.80} & \textbf{63.62} & \textbf{58.14} \\
        \midrule
        VLTSeg**~\cite{hummer2023vltseg} & EVA02-L & 65.60 & 58.40 & 66.50 & 63.50 \\
        Rein**~\cite{wei2024stronger} & EVA02-L & 65.30 & \textbf{60.50} & 64.90 & 63.60 \\
        Rein$\dagger$ & ViT-L & 66.40 & 60.40 & 66.10 & 64.30 \\
        TQDM**~\cite{pak2024textual} & EVA02-L & \underline{68.88} & 59.18 & \underline{70.10} & \underline{66.05} \\
        \rowcolor{gray!40} DPMFormer** (ours) & EVA02-L & \textbf{70.08} & \underline{60.48} & \textbf{70.66} & \textbf{67.07} \\
        \bottomrule
    \end{tabular}
    }
    \caption{Comparison with the state-of-the-art DGSS methods on \textit{synthetic-to-real} scenario with GTAV~\cite{richter2016playing} as a source. We note that *, **, and $\dagger$ denotes the models initialized with pretrained CLIP~\cite{radford2021learning}, EVA02-CLIP~\cite{sun2023eva}, and DINOv2~\cite{oquab2023dinov2}, respectively. The best and the second best performances are highlighted with \textbf{bold} and \underline{underline}, respectively.}
    \label{tab:sota_gta2others}
    
\end{table}

\begin{table}[t]
    \centering
    \resizebox{1.0\linewidth}{!}{
    \begin{tabular}{l l c c c c}
        \toprule
        \textbf{Models (Synthia)} & \textbf{Backbone} & \textbf{Cityscapes} & \textbf{BDD} & \textbf{Mapillary} & \textbf{Avg.} \\
        \midrule
        SAN-SAW~\cite{peng2022semantic} & ResNet-101 & 40.87 & 35.98 & 37.26 & 38.04 \\
        TLDR~\cite{kim2023texture} & ResNet-101 & 42.60 & 35.46 & 37.46 & 38.51 \\
        IBAFormer~\cite{sun2023ibaformer} & MiT-B5 & 50.92 & 44.66 & 50.58 & 48.72 \\
        VLTSeg**~\cite{hummer2023vltseg} & EVA02-L & 56.80 & 50.50 & 54.50 & 53.93 \\
        TQDM**~\cite{pak2024textual} & EVA02-L & \underline{57.99} & \underline{52.43} & \underline{54.87} & \underline{55.10} \\
        \rowcolor{gray!40} DPMFormer** (ours) & EVA02-L & \textbf{58.92} & \textbf{54.39} & \textbf{60.08} & \textbf{57.80} \\
        \bottomrule
    \end{tabular}
    }
    \caption{Comparison with the state-of-the-art DGSS methods where SYNTHIA~\cite{ros2016synthia} is set as a source dataset.}
    \label{tab:sota_syn2others}
    
\end{table}

\subsection{Quantitative Results}

\noindent\textbf{Synthetic-to-real.} As shown in Tab.~\ref{tab:sota_gta2others}, we achieve state-of-the-art in every target domain with both backbones. Notably, with the ViT-B backbone initialized with pretrained CLIP~\cite{radford2021learning}, DPMFormer consistently surpasses previous state-of-the-art~\cite{pak2024textual} by 3.17\% on average mIoU among target domains. In detail, on Cityscapes which contains mostly daytime real images, the domain robustness empowered by consistency learning assists DPMFormer to cope with the domain gap caused by visual realism, improving the state-of-the-art performance by 1.5\%. Meanwhile on BDD100K and Mapillary which have higher environmental variation in terms of weather, time and location, the domain-aware prompt generation enables the model to  adaptively leverage the textual knowledge for segmentation, impressively escalating the performance by 4.14\%, and 3.17\%, respectively. Moreover, even with the larger backbone~\cite{sun2023eva}, we mark the highest average performance of 67.07\%. In Cityscapes and Mapillary dataset, we outperform TQDM~\cite{pak2024textual} by 1.2\% and 0.56\% respectively, while nearly reaching the score of Rein~\cite{wei2024stronger} in BDD. The overall results demonstrate the superiority of DPMFormer, emphasizing the efficacy of domain-awareness as well as domain-robustness.

In Tab.~\ref{tab:sota_syn2others}, we also present synthetic-to-real results where SYNTHIA~\cite{ros2016synthia} is set as a source domain. Again, DPMFormer outperforms all competitors by a large margin, achieving the state-of-the-art performance in every target domain. Remarkably, our framework surpasses the previous state-of-the-art~\cite{pak2024textual} by an average of 2.7\% mIoU across target domains. In particular, the performance on Mapillary~\cite{neuhold2017mapillary} improves significantly by 5.21\%. These results demonstrate the effectiveness of DPMFormer in addressing the domain gap in terms of texture and perspective.

\begin{table}[t]
    \centering
    \resizebox{1.0\linewidth}{!}{
    \begin{tabular}{lcccc}
        \toprule
        \textbf{Models (Cityscapes)} & \textbf{Backbone} & \textbf{BDD} & \textbf{Mapillary} & \textbf{Avg.} \\
        \midrule
        SAN-SAW~\cite{peng2022semantic} & ResNet-101 & 54.73 & 61.27 & 42.43 \\
        WildNet~\cite{lee2022wildnet} & ResNet-101 & 47.01 & 41.73 & 44.87 \\
        SHADE~\cite{zhao2022style} & ResNet-101 & 50.95 & 43.66 & 45.27 \\
        TQDM*~\cite{pak2024textual} & ViT-B & 50.54 & 65.74 & 58.14 \\
        \rowcolor{gray!40}DPMFormer* (ours) & ViT-B & \textbf{54.81} & \textbf{67.72} & \textbf{61.27} \\
        \midrule
        HGFormer~\cite{ding2023hgformer} & EVA02-L & 61.50 & 72.10 & 66.80 \\
        VLTSeg**~\cite{hummer2023vltseg} & EVA02-L & 64.40 & \underline{76.40} & 70.40 \\
        Rein**~\cite{wei2024stronger} & EVA02-L & 64.10 & 69.50 & 66.80 \\
        Rein$\dagger$ & ViT-L & \textbf{65.00} & 72.30 & 68.65 \\
        TQDM~\cite{pak2024textual}** & EVA02-L & \underline{64.72} & 76.15 & \underline{70.44} \\
        \rowcolor{gray!40}DPMFormer** (ours) & EVA02-L & 64.2 & \textbf{76.67} & \textbf{70.44} \\
        \bottomrule
    \end{tabular}
    }
    \caption{Comparison with the state-of-the-arts trained with Cityscapes~\cite{cordts2016cityscapes} on the \textit{real-to-real} scenario.}
    \label{tab:sota_city2others}
\end{table}

\noindent\textbf{Real-to-real.} Furthermore, as shown in Tab.\ref{tab:sota_city2others}, DPMFormer records the highest average mIoU with both backbones. With the CLIP-pretrained backbone, we significantly boost the state-of-the-art performance by 4.27\% and 1.98\% on BDD and Mapillary, respectively. These results indicate that enhanced domain robustness supports the model in predicting consistently under environmental changes, while the domain-aware prompts facilitate the semantic alignment on the unseen domain. As a result, DPMFormer successfully achieves an average performance gain of 3.13\% over the previous state-of-the-art. Equipped with the EVA-CLIP~\cite{sun2023eva} pretrained backbone, we attain the highest mIoU of 76.67\% on Mapillary, while performing slightly lower on BDD. Overall, the average performance is 70.44\%, which is comparable to TQDM~\cite{pak2024textual}.
% the previous state-of-the-art~\cite{pak2024textual}. 

\begin{figure*}[t]
  \centering
  \includegraphics[clip=true,width=0.96\linewidth]{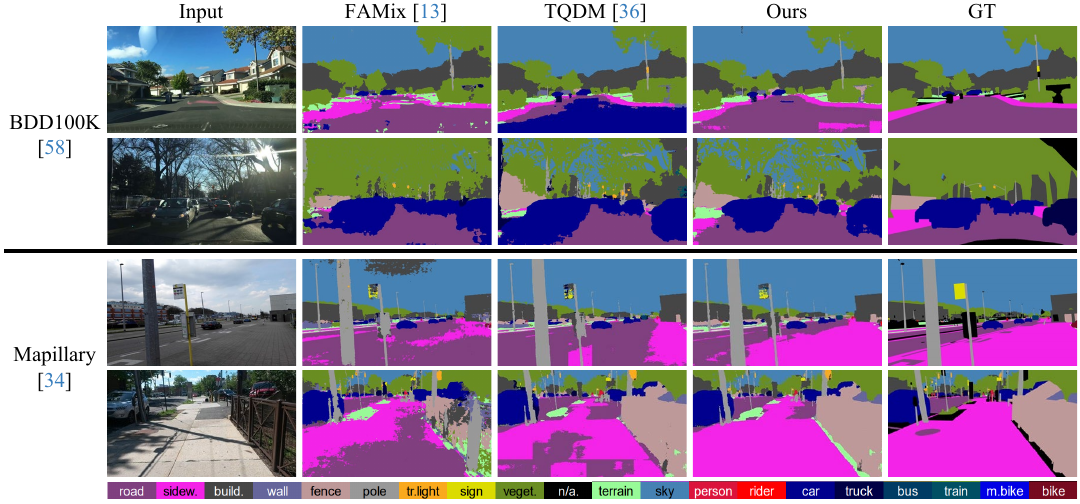}
  \caption{Qualitative comparison on synthetic-to-real scenario with the CLIP-pretrained backbone (ViT-B). The training source domain is set as GTAV~\cite{richter2016playing} while the target domains are BDD100K~\cite{yu2020bdd100k} and Mapillary~\cite{neuhold2017mapillary}. The overall result shows that DPMFormer accomplishes precise segmentation with the images of strong illumination contrast as well as confusing textures.}
  \label{fig:qualitative_comparison}
\end{figure*}

\subsection{Qualitative Results}
In Fig.~\ref{fig:qualitative_comparison}, we qualitatively compare our method with FAMix~\cite{fahes2024simple} and TQDM~\cite{pak2024textual} on the \textit{synthetic-to-real} scenario with the pretrained CLIP backbone (ViT-B). As depicted in the results on BDD100K, both competitors struggle to discriminate accurately under environments having confusing textures or large variations of illumination. In particular, the road in the first-row image has a texture similar to the exterior of a car, leading TQDM to a mismatch between the visual and textual features. In addition, FAMix shows sensitivity to the shades, mislabeling the road as a sidewalk. On the contrary, our method effectively distinguishes the road from other categories by reflecting the domain-specific properties of real-world roads in clear weather. In the case of the second row image with the high illumination contrast, both FAMix and TQDM fail to notice sidewalks in the image due to their low brightness. On the other hand, DPMFormer carries out precise predictions under severe photometric changes owing to the domain robustness acquired from handling diverse texture changes.

As observed with the samples from the Mapillary (the third and fourth rows), FAMix and TQDM lack discriminability between the road and the sidewalk. To be specific with the first image (third row), these classes appear to have a similar color and texture, resulting in misclassifications. Meanwhile, the domain-specific properties from the image transfer the semantic knowledge of the appearance of real-world sidewalk to the model, DPMFormer accurately distinguish between the road and the sidewalk. With the second image, TQDM~\cite{pak2024textual} shows vulnerability against small textural changes on the sidewalk (\eg, shades of the poles). Contrarily, DPMformer effectively performs segmentation owing to its domain-robustness as well as the obtained domain-aware textual queries.

\subsection{Analysis}
\subsubsection{Ablation studies}
To demonstrate the effectiveness of the components in DPMFormer, \ie, texture perturbation, domain-robust consistency learning ($\mathcal{L}_{cons}$), and domain-aware context prompt learning ($\mathcal{L}_{contra}$), we conduct ablation studies with the CLIP pretrained backbone on \textit{synthetic-to-real} scenario. As presented in Tab.~\ref{tab:ablation_study}, every component contributes to the performance gain. Specifically, the texture perturbation enlarges the observable domain during training, boosts the average performance by 0.65\%. In addition with the $\mathcal{L}_{\text{cons}}$, the model successfully equips the robustness against domain-shift and enhances the prediction accuracy on unseen target domains, increasing the average mIoU by 0.77\%. Furthermore, $\mathcal{L}_{\text{contra}}$ allows the model to acquire domain-specific property from the input image in the form of textual prompt embedding. Consequently, the model exploits both visual and textual cues properly aligned for the target domain, considerably elevate the average mIoU by 1.99\%. Combining all components together, DPMFormer effectively learns both domain-awareness and domain-robustness to generate accurate results consistently on unseen target domains.

\begin{table}[t]
    \centering
    \resizebox{\columnwidth}{!}{%
    \begin{tabular}{lcccc}
    \toprule
    \textbf{Models (GTAV)} & \textbf{Cityscapes} & \textbf{BDD} & \textbf{Mapillary} & \textbf{Avg.} \\
    \midrule
    Baseline & 57.5 & 47.66 & 59.76 & 54.97 \\
    + Perturbation & 57.04 & 48.19 & 60.91 & 55.38 \\
    + $\mathcal{L}_{\text{cons}}$ & 58.22 & 49.39 & 60.84 & 56.15 \\
    + $\mathcal{L}_{\text{contra}}$ & \textbf{59.0} & \textbf{51.8} & \textbf{63.62} & \textbf{58.14} \\
    \bottomrule
    \end{tabular}%
    }
    \caption{Ablation study of proposed components. The models are trained on GTAV with CLIP pretrained backbone (ViT-B). The best performance for each column is highlighted in bold.}
    \label{tab:ablation_study}
\end{table}

\subsubsection{Comparison with prompt learning methods}
\label{ablation_cocop}

\begin{figure}[t]
  \centering
  \includegraphics[clip=true,width=\columnwidth]{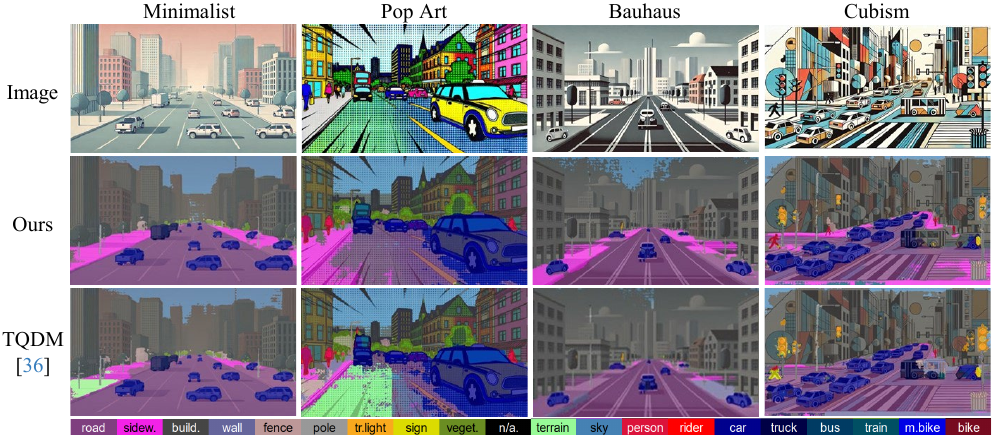}
  \caption{Qualitative results on diverse styles, \ie, \textit{Minimalist}, \textit{Pop Art}, \textit{Bauhaus}, and \textit{Cubism}. The models are trained on GTAV with CLIP pretrained backbone (ViT-B).}
  \label{fig:qualitative_ood}
\end{figure}
% Despite the textural abstractness from artistic expressions, our model achieve accurate segmentation.

In Tab.~\ref{tab:ablation_cocoop}, we compare our domain-aware context prompt learning ($\mathcal{L}_{\text{contra}}$) with previous prompt learning methods~\cite{zhou2022conditional, khattak2023maple, khattak2023self} to validate the efficacy for domain generalization. Given the textual features $t=\textit{ENC}_{T}(p_x)$ and visual features $v=\hat{F}(x)$, the contrastive loss of CoCoOp~\cite{zhou2022conditional} encourages the similarity between the anchor text feature and its corresponding visual feature to be higher than others. As observed in the second row, CoCoOp marginally improves the performance of prompt generator $h_{\theta}$, indicating that the generated feature become instance-specific which is less generalizable to other domains. Moreover, additionally assigning the visual feature from augmented sample as positives (CoCoOp$^+$) resulted in negligible performance gain, since the derived prompt contains domain-invariant information rather than domain-specific properties. Furthermore, although both MaPLe~\cite{khattak2023maple} and PromptSRC~\cite{khattak2023self} employ multi-modal prompts, their performance gains are modest due to the limited generalizability of fixed prompts. On the other hand, our domain-aware contrastive loss design enable the model to proficiently capture domain-specific property from the image which are more helpful for the semantic alignment between the visual features and the textual knowledge. From the fifth to the eighth row, we verify $\mathcal{L}_{\text{contra}}$ with different similarity calculation targets for $\text{sim}(\cdot)$, \ie, text features $t$, text and visual features ($t,v$), and (3) output context embeddings $\pi$. The results confirm that computing the domain-aware contrastive loss empowers the domain generalizability especially when computed with the context embeddings $\pi$ which can provide direct domain guidance to $h_{\theta}$.

\subsubsection{Qualitative results on diverse styles}
To further verify the domain generalization capability of DPMFormer, we compare our model with TQDM~\cite{pak2024textual} on images with diverse artistic styles generated by ChatGPT\footnote{https://chat.openai.com}. As shown in Fig.~\ref{fig:qualitative_ood}, DPMFormer correctly predicts objects and their surroundings even in severe domain shift scenarios, \eg, \textit{Cubism}. With modern art styles (\ie, \textit{Minimalist}, \textit{Pop Art}, and \textit{Bauhaus}), TQDM perplexes among road, sidewalk, and terrain due to their textural changes. Notably, DPMFormer produces more reliable results owing to the enhanced robustness against texture variations that are learned from the texture perturbation and consistency learning. In case of the scene in cubism style, TQDM mispredicts the person on the left side because of its visual similarity with the person in the traffic sign. On the other hand, our method reflects the cubism style to the textual object queries, making an accurate classification with the same instance.

\begin{table}[t]
    \centering
    \resizebox{\columnwidth}{!}{%
    \begin{tabular}{lcccc}
    \toprule
    \textbf{$\mathcal{L}_{\text{contra}}$ (GTAV)} & \textbf{Cityscapes} & \textbf{BDD} & \textbf{Mapillary} & \textbf{Avg.} \\
    \midrule
    -- & 57.65 & 49.63 & 61.10 & 56.13 \\
    CoCoOp~\cite{zhou2022conditional} & 57.84 & 49.91 & 61.33 & 56.36 \\
    $\text{CoCoOp}^{+}$ & 57.60 & 50.04 & 60.96 & 56.20 \\
    MaPLe~\cite{khattak2023maple} & 57.87 & 50.12 & 61.04 & 56.34 \\
    PromptSRC~\cite{khattak2023self} & 58.10 & 49.73 & \underline{62.51} & 56.78 \\
    \midrule
    Ours ($\text{sim}(t,t)$) & 58.17 & 50.18 & 61.37 & 56.57 \\
    Ours ($\text{sim}(t,v)$) & \textbf{59.49} & \underline{50.34} & 62.18 & \underline{57.34} \\
    Ours ($\text{sim}(\pi,\pi)$) & \underline{59.00} & \textbf{51.80} & \textbf{63.62} & \textbf{58.14} \\
    \bottomrule
    \end{tabular}%
    }
    \caption{Comparison with prompt learning methods~\cite{zhou2022conditional, khattak2023maple, khattak2023self} in synthetic-to-real scenario. -- denotes DPMFormer without $\mathcal{L}_{\text{contra}}$. $\text{CoCoOp}^{+}$ indicates the modified loss that additionally include the feature from augmented sample of anchor as positives. $t$, $v$, and $\pi$ refers to the text feature, visual feature, and context embedding.}
    \label{tab:ablation_cocoop}
\end{table}

% \subsection{Visualization}

\section{Conclusion}

% We have presented DPMFormer, a novel framework that addresses the limitations of existing Vision-Language Model-based Domain Generalized Semantic Segmentation (DGSS) methods in fully utilizing textual knowledge for improved domain generalizability. Our approach leverages domain-aware prompt learning and domain-aware contrastive learning to translate input image features into domain-specific textual context prompts, enabling the model to adaptively identify target classes across varying domains. By incorporating textural augmentation and enforcing class and mask consistency losses, we enhance the model's robustness to domain shifts, ensuring consistent and accurate segmentation even under severe textural changes.

In this paper, we presented DPMFormer, a novel domain generalization framework for semantic segmentation. To address the limited generalizability of fixed context prompt learned from a single source dataset, we introduced a novel domain-aware prompt learning which can reflect domain-specific properties of the input image into textual prompts. The proposed component enhanced the semantic alignment between visual and textual cues, assisting the model to fully leverage the abundant semantic knowledge of VLMs. Moreover, to empower domain-robustness, we simulated various domain shifts via texture perturbations, and provided consistency guidance to the model by minimizing prediction discrepancies between original and augmented images. Through extensive experiments and analyses, we demonstrated the effectiveness of DPMFormer, achieving state-of-the-art performance on various benchmarks. 
% Moreover, we verified the efficacy of each proposed components through ablation studies and analyses.

\noindent\textbf{Acknowledgments.} 
{\small This project was partly supported by the National Research Foundation of Korea
 grant funded by the Korea government (MSIT) (No. 2022R1A2B5B02001467; RS-2024-00346364) and the National Research Foundation of Korea (NRF) grant funded by the Korea government (MSIT) (IRIS RS-2025-24803127). This research was supported by Sookmyung Women’s University Research Grants (1-2403-2034).}

{
    \small
    \bibliographystyle{ieeenat_fullname}
    \bibliography{main}
}

\clearpage
\setcounter{page}{1}
\maketitlesupplementary

\section{Details of the Baseline Losses}
\label{sec:suppl_baseline_losses}
Our baseline~\cite{pak2024textual} exploits two type of losses: $\mathcal{L}_{seg}$ for learning semantic segmentation task learning, $\mathcal{L}_{reg}$ for regularization to maintain visual and textual knowledge of the pretrained model.
Specifically, $\mathcal{L}_{seg}$ is formulated as follows:
\begin{equation}
    \mathcal{L}_{seg} = \mathcal{L}_{cls} + \lambda_{bce}\mathcal{L}_{bce} + \lambda_{dice}\mathcal{L}_{dice},
\end{equation}
where $\lambda_{bce}$ and $\lambda_{dice}$ are weight coefficients of their corresponding losses. $\mathcal{L}_{cls}$ is a classification loss for the class predictions $\hat{c}_{q}$, and both $\mathcal{L}_{bce}$ and $\mathcal{L}_{dice}$ losses are binary cross-entropy loss and dice loss for the mask predictions $\hat{y}^{\text{mask}}$. The outputs of each queries are matched to the ground truth class and mask through the fixed matching.

In addition, $\mathcal{L}_{reg}$ is computed as follows:
\begin{equation}
    \mathcal{L}_{reg} = \mathcal{L}^{L}_{reg} + \mathcal{L}^{VL}_{reg} + \mathcal{L}^{V}_{reg},
\end{equation}
where $\mathcal{L}^{L}_{reg}$, $\mathcal{L}^{VL}_{reg}$, and $\mathcal{L}^{V}_{reg}$ refers to language regularization, vision-language regularization, and vision regularization, respectively. Each loss is derived as follows:
\begin{align}
    \mathcal{L}^{L}_{reg} &= \text{Cross-Entropy}(\text{Softmax}(\hat{t}\hat{T}_{0}^{\top}, I_K)), \\
    \mathcal{L}^{VL}_{reg} &= \text{Cross-Entropy}(\text{Softmax}(S/\tau), y), \\
    \mathcal{L}^{V}_{reg} &= \lvert v^{\text{CLS}} - v_0^{\text{CLS}} \rVert_2.
\end{align}
Specifically, $\mathcal{L}^{L}_{reg}$ encourages the text feature $t$ to follow a text feature $T_0$ effective for semantic segmentation task which is obtained with the fixed prompt template `a clean origami of a \{$\text{class}_{k}$\}'~\cite{lin2023clip}. The loss matches the cosine-similarity matrix $\hat{t}\hat{T}_{0}^{\top}$ with the $K$-dimensional identity matrix $I_K$ via the cross-entropy loss. Secondly, $\mathcal{L}^{VL}_{reg}$ enhances the alignment of the visual feature $v = ENC_{I}(x)$ and text feature $t$ by matching the score map $S = \hat{v}\hat{t}^{\top}$ with the ground-truth segmentation map $y$. $\hat{v}$ indicates the normalized visual feature and $\tau$ denotes a temperature coefficient. Lastly, $\mathcal{L}^{V}_{reg}$ contributes to preserving visual knowledge of the VLM during training by minimizing the discrepancy between class tokens $v^{\text{CLS}}$ and $v_0^{\text{CLS}}$ which are obtained from the training backbone and the frozen one, respectively.

Notably, our proposed domain-aware context prompt learning and domain-robust consistency learning are effectively combined with the baseline objectives, significantly improving the overall performance for DGSS task.

\section{Hyperparameter Analysis}
\label{sec:hyperparam}
\begin{table}[h]
    \centering
    \resizebox{0.95\columnwidth}{!}{
    \begin{tabular}{l c c c c c}
        \toprule
        \textbf{Models (GTAV)} & \textbf{Parameter} & \textbf{Cityscapes} & \textbf{BDD} & \textbf{Mapillary} & \textbf{Avg.} \\
        \midrule
        Baseline & - & 57.5 & 47.66 & 59.76 & 54.97 \\
        \midrule
        \multirow{12}{*}{DPMFormer} & $\lambda_{\text{contra}}=0.1$ & 57.95 & 49.97 & 61.03 & 56.32 \\
        & $\lambda_{\text{contra}}=0.5$ & 58.22 & 50.00 & 61.00 & 56.41 \\
        & \cellcolor{lightgray}$\lambda_{\text{contra}}=1$ & \cellcolor{lightgray}59.00 & \cellcolor{lightgray}51.80 & \cellcolor{lightgray}63.62 & \cellcolor{lightgray}58.14 \\
        & $\lambda_{\text{contra}}=10$ & 58.21 & 50.19 & 61.64 & 56.68 \\
        \cmidrule{2-6}
        & $\lambda_{\text{cons}}=1$ & 57.54 & 48.91 & 60.94 & 55.80 \\
        & $\lambda_{\text{cons}}=5$ & 58.10 & 49.48 & 61.20 & 56.26 \\
        & \cellcolor{lightgray}$\lambda_{\text{cons}}=10$ & \cellcolor{lightgray}59.00 & \cellcolor{lightgray}51.80 & \cellcolor{lightgray}63.62 & \cellcolor{lightgray}58.14 \\
        & $\lambda_{\text{cons}}=50$ & 59.08 & 49.82 & 62.03 & 56.98 \\
        \cmidrule{2-6}
        & $\tau=0.1$ & 58.50 & 50.86 & 62.53 & 57.30 \\
        & \cellcolor{lightgray}$\tau=0.5$ & \cellcolor{lightgray}59.00 & \cellcolor{lightgray}51.80 & \cellcolor{lightgray}63.62 & \cellcolor{lightgray}58.14 \\
        & $\tau=1$ & 58.58 & 50.05 & 61.94 & 56.86 \\
        & $\tau=2$ & 56.81 & 49.52 & 60.83 & 55.72 \\
        \bottomrule
    \end{tabular}
    }
    \caption{Hyperparameter analysis on synthetic-to-real scenarios with CLIP backbone (ViT-B).}
    \label{tab:rebuttal_hyperparam}
\end{table}
The quantitative analyses on hyperparameters {$\lambda_{\text{contra}}$, $\lambda_{\text{cons}}$, $\lambda_{\text{tau}}$} are provided in Tab.~\ref{tab:rebuttal_hyperparam}, which are weighting factors of $\mathcal{L}_{\text{contra}}$, $\mathcal{L}_{\text{cons}}$, and a temperature scaler in $\mathcal{L}_{\text{contra}}$.
We note that we empirically set these hyperparameters for the balanced optimization of all training losses. The best performance is obtained when $\lambda_{\text{contra}}$, $\lambda_{\text{cons}}$, $\lambda_{\text{tau}}$ are set as 1.0, 5.0, 0.5, respectively.
Excessively reducing or increasing the weighting factors resulted in marginal improvements over the baseline. The temperature parameter $\tau$ achieved optimal performance at 0.5, adequately reducing the entropy of output distribution in the similarity matrix, thereby facilitating loss convergence.

\section{Model Performance on Diverse Corruptions}
\label{sec:suppl_cityc}
\begin{table}[h]
    \centering
    \resizebox{1.0\columnwidth}{!}{
    \begin{tabular}{lcccccc}
        \toprule
        Method & Blur & Noise & Digital & Weather & Elastic Transform & Average \\
        \midrule
        TQDM~\cite{pak2024textual} & 39.40 & \textbf{20.24} & 52.56 & 46.03 & \textbf{73.50} & 42.02 \\
        DPMFormer & \textbf{40.08} & 18.75 & \textbf{53.57} & \textbf{48.85} & 73.04 & \textbf{42.72} \\
        \bottomrule
    \end{tabular}
    }
    \vspace{-2mm}
\caption{Quantitative evaluation on Cityscapes-to-Cityscapes-C with corruption level 5.}
\label{tab:rebuttal_cityc}
\end{table}

% \begin{table}[h]
%     \centering
%     \resizebox{1.0\columnwidth}{!}{
%     \begin{tabular}{l|cccc|cccc}
%         \toprule
%         & \multicolumn{4}{c|}{Blur} & \multicolumn{4}{c}{Noise} \\
%         \cmidrule(lr){2-5} \cmidrule(lr){6-9}
%         Method & Motion & Defocus & Glass & Gaussian & Gaussian & Impulse & Shot & Speckle \\
%         \midrule
%         Baseline & \textbf{53.77} & \textbf{53.87} & 27.3 & 38.41 & \textbf{14.56} & \textbf{15.18} & \textbf{16.27} & 34.94 \\
%         Ours & 52.95 & 51.62 & \textbf{29.38} & \textbf{41.39} & 11.4 & 13.51 & 13.86 & \textbf{36.24} \\
%         \bottomrule
%     \end{tabular}
%     }
%     \vspace{-2mm}
% \caption{Quantitative evaluation on Cityscapes-to-Cityscapes-C with corruption level 5.}
% \label{tab:rebuttal_cityc}
% \end{table}

% \begin{table}[h]
%     \centering
%     \resizebox{1.0\columnwidth}{!}{
%     \begin{tabular}{l|cccc|cccc}
%         \toprule
%         & \multicolumn{4}{c|}{Digital} & \multicolumn{4}{c}{Weather} \\
%         \cmidrule(lr){2-5} \cmidrule(lr){6-9}
%         Method & Brightness & Contrast & Saturate & JPEG & Snow & Spatter & Fog & Frost \\
%         \midrule
%         Baseline & 73.28 & 46.47 & 69.67 & 36.52 & 34.98 & 50.64 & \textbf{65.32} & 33.18 \\
%         Ours & \textbf{74.00} & \textbf{48.00} & \textbf{70.98} & \textbf{38.93} & \textbf{40.63} & \textbf{54.06} & 64.96 & \textbf{35.73} \\
%         \bottomrule
%     \end{tabular}
%     }
%     \vspace{-2mm}
% \caption{Quantitative evaluation on Cityscapes-to-Cityscapes-C with corruption level 5.}
% \label{tab:rebuttal_cityc}
% \end{table}
We present the model performance on Cityscapes~\cite{cordts2016cityscapes}-to-Cityscapes-C~\cite{hendrycks2019benchmarking} with corruption level 5 in Tab.~\ref{tab:rebuttal_cityc} with the CLIP pretrained backbone (ViT-B). We group corruptions into Blur, Noise, Digital, Weather, and Elastic Transform. As described, DPMFormer surpasses another language-driven DGSS method~\cite{pak2024textual} especially against blur, digital, and weather corruptions that induce a large texture changes.

\section{Domain Generalization for Image Classification}
\label{sec:suppl_dg}
\begin{table}[h]
    \centering
    \resizebox{0.98\columnwidth}{!}{
    \begin{tabular}{lcccc}
        \toprule
        Method & PACS & VLCS & Office-Home & Terra \\
        \midrule
        ZS-CLIP~\cite{radford2021learning} & 90.7 $\pm$ 0.0 & 80.0 $\pm$ 0.0 & 70.8 $\pm$ 0.0 & 23.8 $\pm$ 0.0 \\
        CoCoOp~\cite{zhou2022conditional} & \underline{91.9 $\pm$ 0.6} & \underline{81.8 $\pm$ 0.3} & 73.4 $\pm$ 0.4 & 34.1 $\pm$ 3.0 \\
        DPL~\cite{zhang2023domain} & 91.8 $\pm$ 0.7 & 80.8 $\pm$ 0.8 & 73.6 $\pm$ 0.4 & 34.4 $\pm$ 1.0 \\
        SPG~\cite{bai2025soft} & \textbf{92.8 $\pm$ 0.2} & \textbf{84.0 $\pm$ 1.1} & \underline{73.8 $\pm$ 0.5} & \textbf{37.5 $\pm$ 1.8} \\
        DPMFormer & 91.5 $\pm$ 0.3 & 81.5 $\pm$ 1.0 & \textbf{73.9 $\pm$ 0.4} & \underline{35.0 $\pm$ 2.1} \\
        \bottomrule
    \end{tabular}
    }
\vspace{-2mm}
\caption{Comparisons on image classification DG methods.}
\label{tab:rebuttal_classification}
\end{table}
In Tab.~\ref{tab:rebuttal_classification}, we evaluate DPMFormer on multi-source domain generalization benchmarks~\cite{Li2017DeeperBA, torralba2011unbiased, venkateswara2017deep, beery2018recognition} with CLIP ResNet50 backbone. Following conventions, the evaluation is conducted in the leave-one-domain-out manner and we report the average domain accuracy of the model selected using the training-domain validation set method. In summary, DPMFormer achieves performance comparable to previous prompt learning methods for image classification~\cite{zhang2023domain, bai2025soft}. In particular, we achieve the best performance on Office-Home~\cite{venkateswara2017deep}, and surpass CoCoOp~\cite{zhou2022conditional} and DPL~\cite{bai2025soft} on Terra-Incognita~\cite{beery2018recognition}.

We note that existing prompt learning studies for image classification~\cite{zhang2023domain, bai2025soft} are not suitable for the single-source setting of DGSS, as they either require multiple source datasets~\cite{zhang2023domain} or depend on multi-stage training and adversarial learning~\cite{bai2025soft} to obtain prompts. In contrast, our domain-aware prompt generation requires image transformations and a contrastive objective, making it more effective for semantic segmentation tasks.

\section{Computational Overhead}
With DPMFormer, each training iteration takes 1.712 seconds, slightly more than the baseline’s 1.501 seconds. Meanwhile, its inference time of 1.376 seconds per batch remains comparable to 1.004 seconds of the baseline. The batch size is reduced by half due to texture perturbation, but still maintains better performance under the same training setting.

\section{Class-wise Quantitative Comparison}
Through Tab.~\ref{tab:suppl_class_g2c} to Tab.~\ref{tab:suppl_class_c2m}, we compare class-wise IoU of DPMFormer with TQDM~\cite{pak2024textual} with the CLIP initialized models. Noticeably, DPMFormer demonstrates higher performance in most classes in various scenarios and shows comparable score even in other cases. In summary, the average IoU consistently outperforms the competitor, verifying the superiority of DPMFormer in DGSS.
% G2C Table
\begin{table*}[ht]
\centering
\resizebox{1.0\linewidth}{!}{
\begin{tabular}{lcccccccccccccccccccc}
\toprule
Method & Road & Sidewalk & Building & Wall & Fence & Pole & Traffic Light & Traffic Sign & Vegetation & Terrain & Sky & Person & Rider & Car & Truck & Bus & Train & Motorcycle & Bicycle & Avg \\
\midrule
TQDM & \textbf{90.97} & \textbf{50.91} & \textbf{88.24} & 36.52 & 38.54 & \textbf{47.76} & 54.82 & 45.78 & 89.10 & \textbf{40.78} & \textbf{89.67} & 74.33 & 40.46 & 85.73 & 39.41 & \textbf{60.69} & \textbf{46.71} & 29.85 & 47.84 & 57.79 \\
Ours & 87.92 & 46.60 & 88.19 & \textbf{38.83} & \textbf{39.37} & 47.27 & \textbf{54.90} & \textbf{49.24} & \textbf{89.11} & 40.49 & 89.52 & \textbf{74.90} & \textbf{43.08} & \textbf{88.11} & \textbf{54.42} & 55.95 & 35.72 & \textbf{44.16} & \textbf{53.15} & \textbf{59.00} \\
\bottomrule
\end{tabular}
}
\caption{Class-wise quantitative comparison (IoU) in synthetic-to-real (GTA~\cite{richter2016playing}-to-Cityscapes~\cite{cordts2016cityscapes}) scenario with the CLIP-pretrained ViT-B backbone.}
\label{tab:suppl_class_g2c}
\end{table*}

% G2B Table
\begin{table*}[ht]
\centering
\resizebox{1.0\linewidth}{!}{
\begin{tabular}{lcccccccccccccccccccc}
\toprule
Method & Road & Sidewalk & Building & Wall & Fence & Pole & Traffic Light & Traffic Sign & Vegetation & Terrain & Sky & Person & Rider & Car & Truck & Bus & Train & Motorcycle & Bicycle & Avg \\
\midrule
TQDM & 88.76 & 48.75 & 79.85 & 22.46 & 30.48 & \textbf{41.94} & 45.72 & 39.35 & 75.04 & 40.69 & 88.11 & 58.43 & 26.54 & 80.00 & 32.39 & 43.10 & 0.00 & 44.99 & 33.26 & 48.41 \\
Ours & \textbf{90.79} & \textbf{49.72} & \textbf{81.38} & \textbf{29.56} & \textbf{34.65} & 41.68 & \textbf{47.03} & \textbf{42.66} & \textbf{75.82} & \textbf{42.24} & \textbf{88.30} & \textbf{59.72} & \textbf{32.23} & \textbf{84.04} & \textbf{37.49} & \textbf{61.20} & 0.00 & \textbf{50.12} & \textbf{35.61} & \textbf{51.80} \\
\bottomrule
\end{tabular}
}
\caption{Class-wise quantitative comparison (IoU) in synthetic-to-real (GTA~\cite{richter2016playing}-to-BDD~\cite{yu2020bdd100k}) scenario with the CLIP-pretrained ViT-B backbone.}
\label{tab:suppl_class_g2b}
\end{table*}

% G2M Table
\begin{table*}[ht]
\centering
\resizebox{1.0\linewidth}{!}{
\begin{tabular}{lcccccccccccccccccccc}
\toprule
Method & Road & Sidewalk & Building & Wall & Fence & Pole & Traffic Light & Traffic Sign & Vegetation & Terrain & Sky & Person & Rider & Car & Truck & Bus & Train & Motorcycle & Bicycle & Avg \\
\midrule
TQDM & 89.93 & 54.28 & 85.24 & 41.74 & 43.44 & \textbf{51.80} & \textbf{56.93} & 67.24 & 79.36 & 50.58 & 94.27 & \textbf{75.94} & \textbf{56.34} & 86.62 & 51.80 & 54.65 & 19.76 & 57.79 & 41.03 & 60.99 \\
Ours & \textbf{90.20} & \textbf{58.33} & \textbf{85.42} & \textbf{43.19} & \textbf{45.21} & 51.57 & 56.36 & \textbf{68.49} & \textbf{79.91} & \textbf{51.64} & \textbf{94.39} & 75.28 & 56.10 & \textbf{88.70} & \textbf{59.99} & \textbf{61.80} & \textbf{32.77} & \textbf{62.89} & \textbf{46.65} & \textbf{63.35} \\
\bottomrule
\end{tabular}
}
\caption{Class-wise quantitative comparison (IoU) in synthetic-to-real (GTA~\cite{richter2016playing}-to-Mapillary~\cite{neuhold2017mapillary}) scenario with the CLIP-pretrained ViT-B backbone.}
\label{tab:suppl_class_g2m}
\end{table*}

% C2B Table
\begin{table*}[h]
\centering
\resizebox{1.0\linewidth}{!}{
\begin{tabular}{lcccccccccccccccccccc}
\toprule
Method & Road & Sidewalk & Building & Wall & Fence & Pole & Traffic Light & Traffic Sign & Vegetation & Terrain & Sky & Person & Rider & Car & Truck & Bus & Train & Motorcycle & Bicycle & Avg \\
\midrule
TQDM & 92.55 & 56.65 & 83.66 & 24.37 & 33.93 & 42.44 & 46.87 & 48.22 & \textbf{84.02} & \textbf{45.68} & \textbf{93.80} & 58.51 & 28.42 & 86.90 & 38.87 & 38.8 & 0.27 & 37.23 & 26.90 & 50.95 \\
Ours & \textbf{92.76} & \textbf{57.68} & \textbf{83.83} & \textbf{28.73} & \textbf{39.57} & \textbf{45.04} & \textbf{49.91} & \textbf{51.20} & 83.96 & 44.53 & 93.78 & \textbf{62.57} & \textbf{40.21} & \textbf{87.37} & \textbf{40.15} & \textbf{47.95} & \textbf{0.29} & \textbf{53.07} & \textbf{38.78} & \textbf{54.81} \\
\bottomrule
\end{tabular}
}
\caption{Class-wise quantitative comparison (IoU) in real-to-real (Cityscapes~\cite{cordts2016cityscapes}-to-BDD~\cite{yu2020bdd100k}) scenario with the CLIP-pretrained ViT-B backbone.}
\label{tab:suppl_class_c2b}
\end{table*}

% G2M Table
\begin{table*}[h]
\centering
\resizebox{1.0\linewidth}{!}{
\begin{tabular}{lcccccccccccccccccccc}
\toprule
Method & Road & Sidewalk & Building & Wall & Fence & Pole & Traffic Light & Traffic Sign & Vegetation & Terrain & Sky & Person & Rider & Car & Truck & Bus & Train & Motorcycle & Bicycle & Avg \\
\midrule
TQDM & \textbf{90.66} & \textbf{53.19} & \textbf{87.34} & 48.39 & 54.05 & 51.64 & 58.57 & 73.51 & 83.93 & 53.16 & 94.64 & 74.39 & 61.32 & 89.62 & 58.58 & 56.80 & 21.92 & 58.84 & 56.87 & 64.60 \\
Ours & 90.65 & 52.88 & 86.99 & \textbf{48.84} & \textbf{57.20} & \textbf{54.03} & \textbf{61.57} & \textbf{76.29} & \textbf{88.21} & \textbf{53.30} & \textbf{97.10} & \textbf{77.04} & \textbf{65.38} & \textbf{90.45} & \textbf{62.03} & \textbf{66.30} & \textbf{24.87} & \textbf{68.04} & \textbf{65.51} & \textbf{67.72} \\
\bottomrule
\end{tabular}
}
\caption{Class-wise quantitative comparison (IoU) in real-to-real (Cityscapes~\cite{cordts2016cityscapes}-to-Mapillary~\cite{neuhold2017mapillary}) scenario with the CLIP-pretrained ViT-B backbone.}
\label{tab:suppl_class_c2m}
\end{table*}

\section{Precision-Recall Curve Comparison}
\begin{figure*}[t]
  \centering
  \includegraphics[clip=true,width=1.0\linewidth]{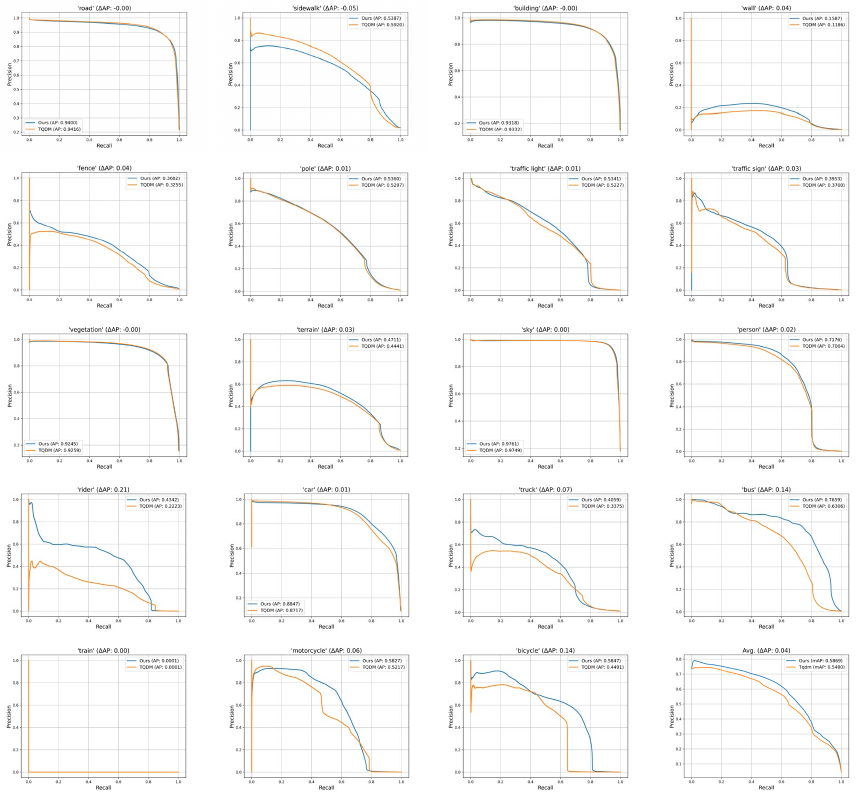}
  \caption{Precision-Recall curve and Average Precision (AP) on synthetic-to-real scenario (GTA~\cite{richter2016playing}-to-BDD~\cite{yu2020bdd100k}) with the CLIP-pretrained backbone (ViT-B).}
  \label{fig:suppl_prcurve_clip}
\end{figure*}

\begin{figure*}[t]
  \centering
  \includegraphics[clip=true,width=1.0\linewidth]{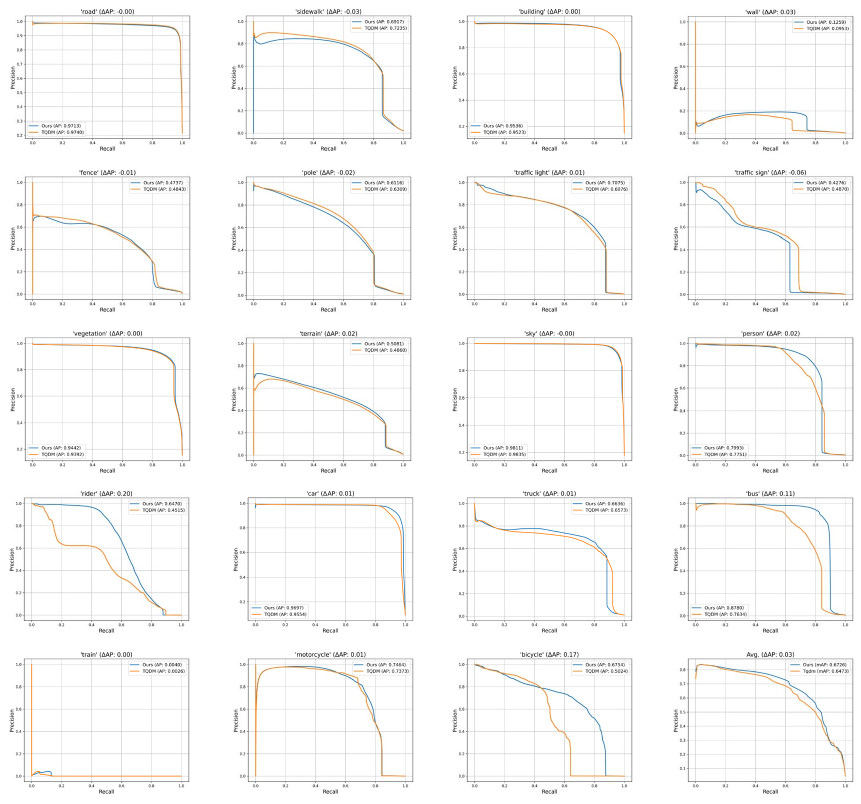}
  \caption{Precision-Recall curve and Average Precision (AP) on synthetic-to-real scenario (GTA~\cite{richter2016playing}-to-BDD~\cite{yu2020bdd100k}) with the EVA02-CLIP~\cite{sun2023eva} pretrained backbone.}
  \label{fig:suppl_prcurve_eva}
\end{figure*}

In Fig.~\ref{fig:suppl_prcurve_clip} and \ref{fig:suppl_prcurve_eva}, we depict Precision-Recall curves of each class with Averaged Precision (AP) in synthetic-to-real scenario (GTA~\cite{richter2016playing}-to-BDD~\cite{yu2020bdd100k}) with CLIP (ViT-B) and EVA02-CLIP backbones, respectively. Compared to TQDM~\cite{pak2024textual}, DPMFormer shows better performance in most classes, validating the effectiveness of domain-aware context prompt learning as well as consistency learning.

\section{Limitation and Future Work}
\label{sec:limitation}
Domain-aware context prompt learning utilizes the global representation of the frozen backbone to obtain domain-specific properties of the image. However, some local textures may differ from the global textures in complex scenes. For example, in night driving scene, the roads are brightened due to car headlights, whereas the sky and surroundings are darkened. Hence, exploiting local texture patterns for more detailed prompt generation can be a good initial motivation for future language-driven DGSS. In addition, the design of the domain-aware prompt generator $h_{\theta}$ and textural perturbations can be further advanced to accomplish better performance. Meanwhile, the unshared label space between the source and the target domain hinders the model from correctly interpreting the image context. From this perspective, we believe addressing DGSS through open-set domain adaptation and the integration of VLM should be a promising direction for future research.

\section{Additional Qualitative Results}
\label{sec:suppl_qualitative}
\begin{figure*}[t]
  \centering
  \includegraphics[clip=true,width=1.0\linewidth]{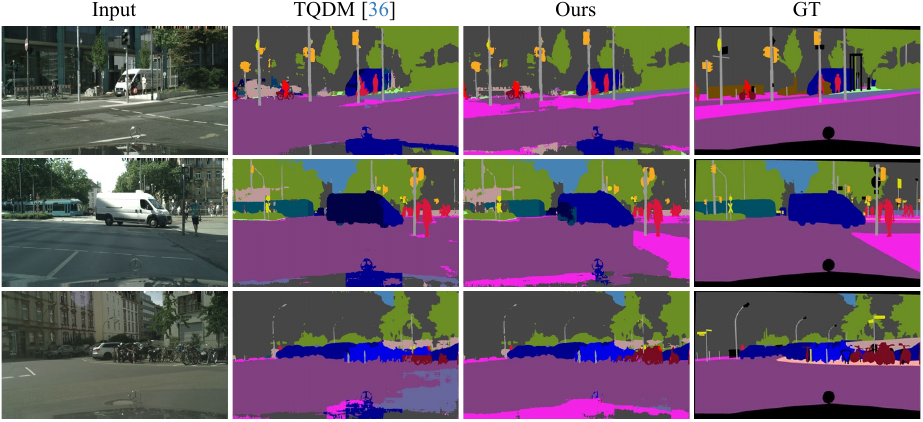}
  \caption{Qualitative comparison on synthetic-to-real scenario (GTA~\cite{richter2016playing}-to-Cityscapes~\cite{cordts2016cityscapes}) with the CLIP-pretrained backbone (ViT-B). With the first image, TQDM~\cite{pak2024textual} mispredicts sidewalk as roads and shows confusion on the region next to the bicycle rider. TQDM also confuses the car with the truck (second row) and misclassify bicycles as motorcycles (third row). On the other hand, DPMFormer produces more reliable and accurate segmentation results in these scenes.}
  \label{fig:suppl_g2c}
\end{figure*}

\begin{figure*}[h]
  \centering
  \includegraphics[clip=true,width=1.0\linewidth]{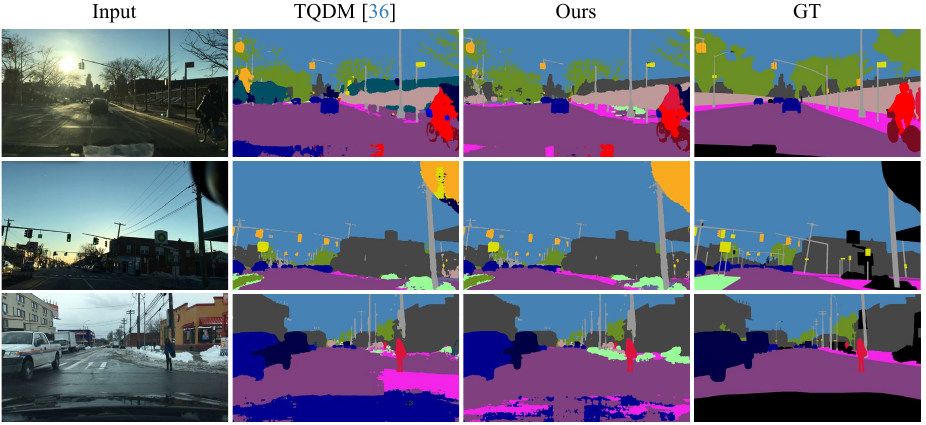}
  \caption{Qualitative comparison on synthetic-to-real scenario (GTA~\cite{richter2016playing}-to-BDD~\cite{yu2020bdd100k}) with the CLIP-pretrained backbone (ViT-B). Due to the large illumination contrast caused from the intense sunlight (first row), TQDM~\cite{pak2024textual} wrongly mark the buliding as a train. In addition, TQDM perplexes the road as `car' and `sidewalk' due to their textural similarity. Contrarily, DPMFormer shows consistent performance under various environments, almost reaching ground-truth segmentation maps.}
  \label{fig:suppl_g2b}
\end{figure*}

\begin{figure*}[h]
  \centering
  \includegraphics[clip=true,width=1.0\linewidth]{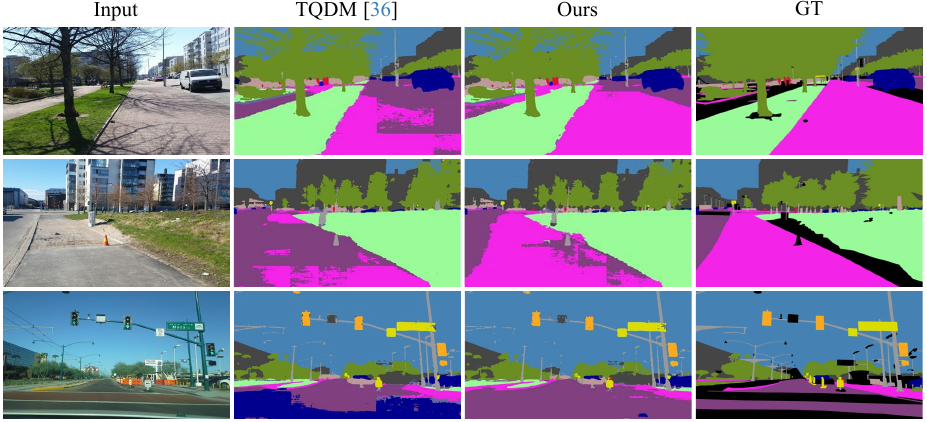}
  \caption{Qualitative comparison on synthetic-to-real scenario (GTA~\cite{richter2016playing}-to-Mapillary~\cite{neuhold2017mapillary}) with the CLIP-pretrained backbone (ViT-B). TQDM~\cite{pak2024textual} confounds road as sidewalk or car because of the textual changes gap from the synthetic texture. Conversely, DPMFormer predicts accurately by utilizing domain-aware context prompt and the domain-robust cues learned from consistency losses.}
  \label{fig:suppl_g2m}
\end{figure*}

\begin{figure*}[t]
  \centering
  \includegraphics[clip=true,width=1.0\linewidth]{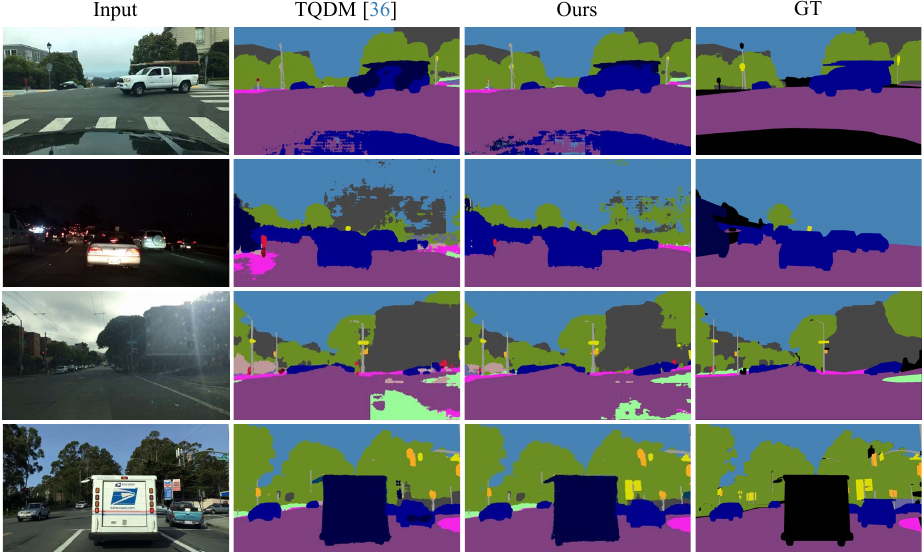}
  \caption{Qualitative comparison on real-to-real scenario (Cityscapes~\cite{cordts2016cityscapes}-to-BDD~\cite{yu2020bdd100k}) with the CLIP-pretrained backbone (ViT-B). In the first image, TQDM mispredicts the car and the bus due to the occlusion. In case of the nighttime (second row) and the daytime (third row) scenes, predictions gets noisy due to the textural ambiguity. As shown in the last row, TQDM fails to catch traffic signs which have different design from the Cityscapes dataset. DPMFormer demonstrates its efficacy by producing more precise segmentation results compared to TQDM.}
  \label{fig:suppl_c2b}
\end{figure*}

\begin{figure*}[h]
  \centering
  \includegraphics[clip=true,width=1.0\linewidth]{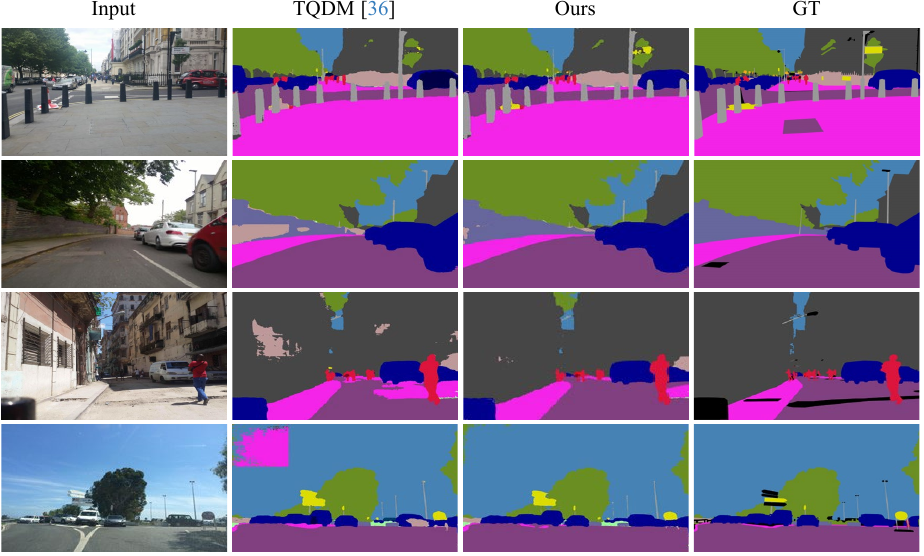}
  \caption{Qualitative comparison on real-to-real scenario (Cityscapes~\cite{cordts2016cityscapes}-to-Mapillary~\cite{neuhold2017mapillary}) with the CLIP-pretrained backbone (ViT-B). Due to the location difference between the datasets, TQDM miss traffic signs (first row) and misclassify the walls (second row) and the road (third row). Also in the clean daytime image (fourth row), fallacious predictions are observed in the sky and infront of the car on the right side. On the other hand, DPMFormer generates clean and reliable predictions among these images.}
  \label{fig:suppl_c2m}
\end{figure*}

Through Fig.~\ref{fig:suppl_g2c} to Fig.~\ref{fig:suppl_c2m}, we provide additional qualitative results of DPMFormer in various scenarios. DPMFormer consistently yields more accurate segmentation results than TQDM~\cite{pak2024textual} in scenes under diverse environments and from various locales.

\end{document}